\documentclass[11pt,a4paper]{article}
\usepackage[hyperref]{acl2020}
\usepackage{times}
\usepackage{latexsym}

\usepackage{arydshln}
\usepackage{graphicx}
\usepackage{subfigure}
\usepackage{url}
\usepackage{multirow}
\usepackage{soul}
\usepackage[ruled,vlined,linesnumbered]{algorithm2e}

\newcommand{\bc}{\mathbf{c}}

\newcommand{\cG}{\mathcal{G}}

\newcommand{\cL}{\mathcal{L}}

\usepackage{microtype}

\aclfinalcopy 

\title{Adv-BERT: BERT is not robust on misspellings! \\Generating nature adversarial samples on BERT}

\author{Lichao Sun$^*$, Kazuma Hashimoto, Wenpeng Yin, \\
\textbf{Akari Asai, Jia Li, Philip Yu$^*$ and Caiming Xiong} \\
University of Illinois at Chicago$^*$ and Salesforce Research \\
  {\tt \small \{james.lichao.sun\}@gmail.com} 
}

\date{}

\begin{document}
\maketitle
\begin{abstract}
    There is an increasing amount of literature that claims the brittleness of deep neural networks in dealing with adversarial examples that are created maliciously. It is unclear, however, how the models will perform in realistic scenarios where  \textit{natural rather than malicious} adversarial instances often exist. This work systematically explores the robustness of BERT, the state-of-the-art Transformer-style model in NLP, in dealing with noisy data, particularly mistakes in typing the keyboard,  that occur inadvertently. Intensive experiments on sentiment analysis and question answering benchmarks indicate that: (i) Typos in various words of a sentence do not influence equally. The typos in informative words make severer damages; (ii) Mistype is the most damaging factor, compared with inserting, deleting, etc.;  (iii) Humans and machines have different focuses on recognizing adversarial attacks.
\end{abstract}

\section{Introduction}

\begin{figure}[tb]
 \setlength{\belowcaptionskip}{-10pt}
 \setlength{\abovecaptionskip}{0pt}
\centering
\includegraphics[width=0.4\textwidth]{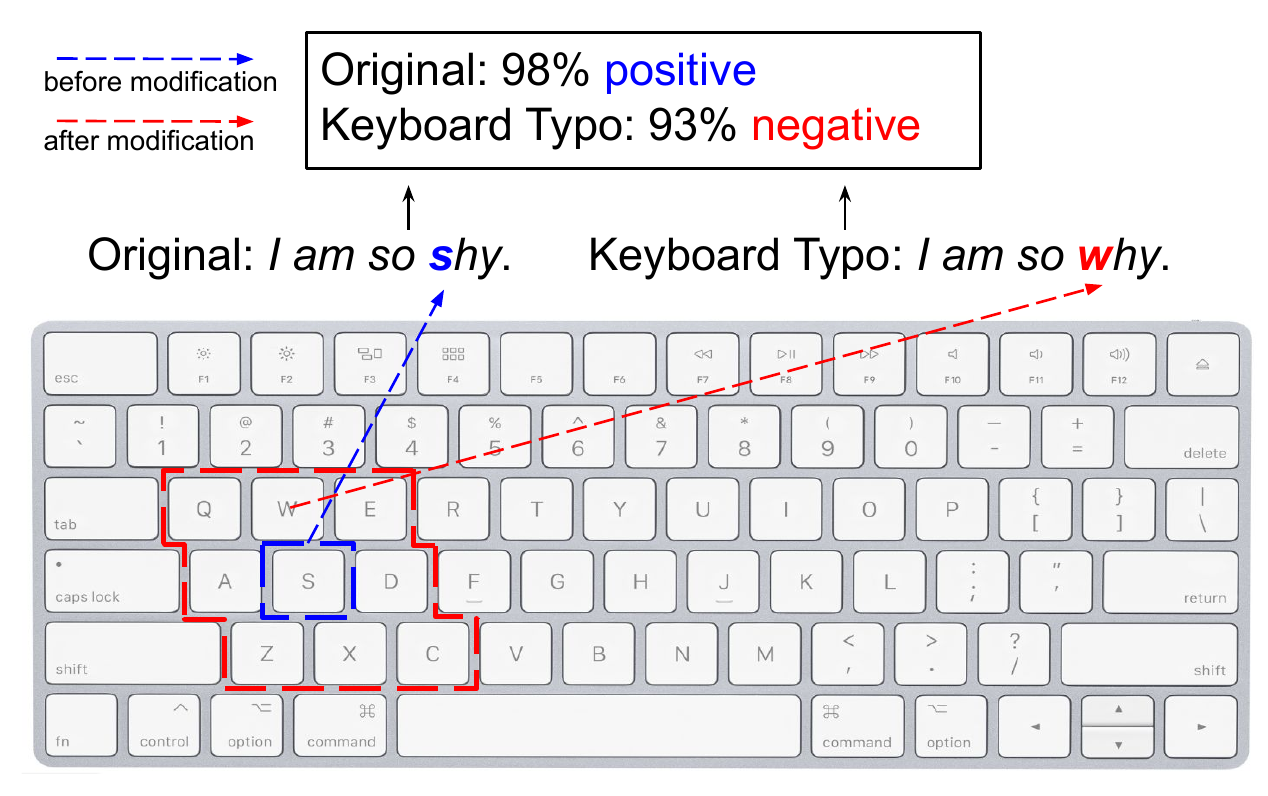}
\caption{Adversarial typo examples for sentiment classification. Only one character substitution will be wrongly classified by BERT-based neural network. Red line indicates the original informative character, and blue line indicate the possible wrongly typing area which is surrounding the legitimate character.} \label{fig:typo}
\end{figure}

Most neural models, such as Recurrent Neural Network \cite{DBLPBahdanauCB14}, Attentive Convolution \cite{DBLPYinS18}, BERT \cite{devlin2018bert}, etc.,  are evaluated on clean  datasets. 
When deploying these models in real-world scenarios, the models have to address  user-generated noisy text.
One of the most common noisy text is  typos because of human mistakes in typing words,
such as character substitution, additional or missing characters.
Even a small typo in Fig. \ref{fig:typo} may confuse the most advanced models like BERT \cite{devlin2018bert},
then a question arises: ``how robust is BERT with respect to keyboard typos?''

Existing approaches generate malicious adversarial examples as attacks by finding the minimum perturbation on each sample to study the robustness of a model \cite{liang2017deep, ebrahimi2017hotflip, gao2018black, li2018textbugger,alzantot2018generating, renetal2019generating}. 
These attacking examples are not that meaningful, therefore can be easily recognized by humans.
In order to prevent humans from identifying the adversarial examples, the community starts to explore more natural perturbations to create the attacks. 
\newcite{zhao2017generating} generates more natural adversarial examples using GANs.  Both \cite{ribeiro2018semantically} and \cite{sato2018interpretable} generate samples with similar semantics. 
\citet{belinkov2017synthetic} presents the first work that studies adversarial typos on the keyboard. 

The character distribution on the keyboard results in a special type of noisy examples, which are natural and unintentional. If conventional attacking examples represent the worse case of inputs, the keyboard-constrained adversarial examples represent the more realistic inputs.
In this work, we systematically study the robustness of the current state-of-the-art neural model BERT in dealing with those inadvertently generated adversarial inputs. Intensive experiments on sentiment analysis and question answering benchmarks indicate that:

\textbullet\enspace BERT has unbalanced attention to the typos in the input. Some typo words have a clear influence on the performance; some, instead, show tiny influence. In addition, different typo generation approaches also show different degrees of damages. Mistype is the severest source of typos;

\textbullet\enspace Machines and humans have different focuses on the typos.  BERT pays more attention to the typos in informative words; in contrast,  humans can better recognize  the typos in some frequent while less informative words;

\textbullet\enspace The robustness of a system is dependent on the learning algorithms as well as the tasks. We found that the BERT-based question answering system on SQUAD \cite{DBLPRajpurkarZLL16} is more brittle than the BERT-based sentiment recognizer. 



This is the first work that systematically studies the robustness of BERT, a Transformer-style \cite{DBLPVaswaniSPUJGKP17} neural model, in addressing noises that appear naturally. Our observations hopefully can provide new insights to the community for building more trustworthy machine intelligence.

\vspace{-5pt}

\section{Generation of natural adversarial examples}

Our adversarial examples are generated with the following principle. We have a pre-trained state-of-the-art model $f: X \rightarrow Y$
for natural language processing tasks,
where $X$ are the text inputs and $Y$ are the corresponding labels.
An adversarial keyboard typo example, denoted as $x_{adv}$, is generated from the original input $x \in X$ and the original prediction $y \in Y$ based on gradient information or random modification.
For each generated example, it will change the model prediction: $f(x_{adv}) = y_{adv} \neq y$.
\begin{algorithm}[tb!]
{\small
\SetAlgoLined
\KwIn{Original document $x$ and its ground truth label $y$, classifier $f(\cdot)$, budget $K$}
\KwOut{True or False}
 initialization $x_{best} \leftarrow x$; $i \leftarrow 0$\; 
 \While{$i < K$}{
 $\bc \leftarrow Segmentation(x_{best}$)\; 
 \For{each component $c_i$ in $\bc$}{
  Compute gradients $\cG$ of component $c_i$ according to Eq.\ref{eq:gradient} \; \label{line:seg}
 }
 Find the component $c$ with largest or smallest gradient norm\;
  $word \leftarrow BackTrackToWord(c)$\; \label{line:word}
  $typos \leftarrow Keyboard\_Typo(word)$\; \label{line:typo}
  $score_h \leftarrow -1$\;
  \For{each $typo$ in $typos$}{
    Generate $x_{typo}$ by replacing $word$ as $typo$ in $x_{best}$\;
    $y', score \leftarrow f(x_{best})$\;
    \If{$y \neq y'$}{
       return True
   }
   \If{$score > score_h$}{
        $x_{best} \leftarrow x_{typo}$
   }
   }
  $i \leftarrow i + 1$;
 }
    Return False
 }
 \caption{Generating Adversarial Samples via Keyboard Typos}
 \label{alg:rekey}
\end{algorithm}
\vspace*{-.2cm}

To start, a sentence $s$ is first tokenized into words (or subwords) $s = (w_1, w_2, \ldots, w_N)$ (line \ref{line:seg} in the Algorithm \ref{alg:rekey}), where $w_i$ is the $i$-th item, and $N$ is the  length after tokenization.
Let $\cL(w, y)$ denote the loss with respect to $w$ under the ground truth label $y$.
Then, we can compute the partial derivative of each item $w_i$ based on the golden output $y$ as shown in this Equation:
\begin{equation}
    \cG_{f}(w_i)  = \bigtriangledown_{w_i} \cL(w_i, y). \label{eq:gradient}
\end{equation}
Based on the gradient information, we can track back to the most informative/uninformative word $x$ though the component $w$ of the word (line \ref{line:word}).
We are interested in generating typos regarding three kinds of words: (1) informative words which have the largest gradient; (2) uninformative words which have the smallest gradient; and (3) random words.

For each word, we consider the following six sources of typos:
(1) \textit{Insertion}: Insert characters or spaces into the word, such as ``oh!'' $\rightarrow$ ``ooh!''.
(2) \textit{Deletion}: Delete a random character of the word because of fast typing, such as ``school'' $\rightarrow$ ``schol''.
(3) \textit{Swap}: Swap random two adjacent characters in a word.
(4) \textit{Mistype}: Mistyping a word though keyboard, such as ``oh'' $\rightarrow$ ``0h''.
(5) \textit{Pronounce}: Wrongly typing due to the close pronounce of the word, such as ``egg'' $\rightarrow$ ``agg''.
(6) \textit{Replace-W}: Replace the word by the frequent human behavioral keyboard typo based on the statistics.\footnote{\url{https://en.wikipedia.org/wiki/Wikipedia:Lists_of_common_misspellings}}

Note that we do not sample characters from random distribution  to implement above modifications, all operations are constrained by the character distribution on the keyboard. The whole generation algorithm is demonstrated in the Algorithm \ref{alg:rekey}. There could be more than one typo in each piece of text through keyboard in real life;  this work  limits the maximal number of typos in one sentence to be $K$.

\vspace{-5pt}
\section{Experiments}

We evaluate the start-of-the-art model BERT\footnote{We use ``bert-base-uncased'' throughout.} on two NLP tasks: sentiment analysis and question answering. On each benchmark, we report the system performance when putting typos in the informative words (i.e., maximal gradient), uninformative words (i.e., minimal gradient) and random words.

\paragraph{Subword tokenization.}
The built-in tokenizer in BERT first performs simple white-space tokenization,  then applies {\it WordPiece} tokenization \cite{DBLPWuSCLNMKCGMKSJL16}. A word can be split into character ngrams (e.g. adversarial$\rightarrow$[ad, \#\#vers, \#\#aria, \#\#l], robustness$\rightarrow$[robust, \#\#ness]).
``\#\#'' is a special symbol to handle the subwords, and we omit it when injecting our typos.
An example typo for ``robustness'' is ``robustnesss'', which is split  into subwords [robust, \#\#ness, \#\#s].

\subsection{Sentiment analysis}

\begin{table}[t]

\begin{center}

\begin{tabular}{r|c|c|c}

$K$ & Max-grad & Min-grad & Random \\ \hline
0 & \multicolumn{3}{c}{92.9} \\ \hdashline
1 & 69.0 & 91.7 & 86.1$\pm$0.8 \\
2 & 62.3 & 90.4 & 80.6$\pm$0.4 \\
3 & 57.5 & 89.0 & 74.8$\pm$1.1 \\
4 & 54.1 & 88.1 & 67.1$\pm$0.8 \\
5 & 53.6 & 87.2 & 60.8$\pm$0.5 \\
6 & 53.1 & 85.7 & 56.8$\pm$0.5 \\
7 & 53.0 & 84.9 & 51.6$\pm$0.5 \\
8 & 53.0 & 84.4 & 47.1$\pm$0.3 \\
9 & 53.0 & 84.1 & 42.2$\pm$0.9 \\
10& 53.0 & 83.9 & 39.3$\pm$1.4 \\ \hline

\end{tabular}

\end{center}
\caption{SST results on the development set. For ``Random'', we show the standard deviations over five different random seeds.}
\label{tab:sst_results_dev}

\end{table}

\begin{table}[t]

\begin{center}

\begin{tabular}{r|c|c|c}


$K$ & Max-grad & Min-grad & Random \\ \hline
1 & 15.9 & 19.8 & 20.0 \\
2 & 15.8 & 28.7 & 33.4 \\ 
3 & 25.1 & 33.4 & 39.9 \\
4 & 20.8 & 38.7 & 44.2 \\
5 & 18.8 & 41.3 & 45.6 \\ \hline

\end{tabular}
\end{center}
\caption{Human capability of recognizing word typos.}
\label{tab:human_read}

\end{table}

We work on the Stanford Sentiment Treebank (SST) \cite{DBLPSocherPWCMNP13} in the binary prediction setting. The standard split has 6920 training, 872 development
and 1821 test sentences. \newcite{DBLPSocherPWCMNP13} used the Stanford Parser
\cite{DBLPKleinM03} to parse each sentence
into subphases. The subphases were then labeled
by human annotators in the same way as the sentences were labeled. Labeled phrases that occur
as subparts of the training sentences are treated
as independent training instances

\paragraph{Results and Analysis.} Table \ref{tab:sst_results_dev} lists the results when we try $K=\{0,1, \cdots, 10\}$ typos in informative words (i.e., ``max-grad''), uninformative words (i.e., ``min-grad'') and random words. The first column (``max-grad'') shows that the model is highly sensitive to the typos on words with the largest gradient norms.
Injecting only a single typo degrades the accuracy by 22.6\%,  suggesting that the gradient norm is a strong indicator to find task-specific informative words.
Another interesting observation is that the accuracy converges to almost the chance-level accuracy (i.e., around 50\%).
In contrast, the second column (``min-grad'') indicates that the model is not sensitive to the typos in words with the smallest gradient norms.
Even with $K=10$, the accuracy drops merely by 9\%, which indicates that in some lucky cases  human keyboard typos may not influence the final predictions.

This comparison discovers the unbalanced attention of BERT to the typos in the input. Only the adversarial attacks based on informative parts really matter. 
However, after $K=7$, any typos in the words with the largest gradient norms can't make the model predictions worse any more, because the current most informative words already contain typos, and then it would not change any words in the sentence.
In this case, if an adversary wants to attack BERT intentionally,
the best strategy is adaptively mixing up ``max-grad'' and ``random'' policy for adversarial sample generation. 

\begin{figure}[tb]
\centering
\includegraphics[width=0.4\textwidth]{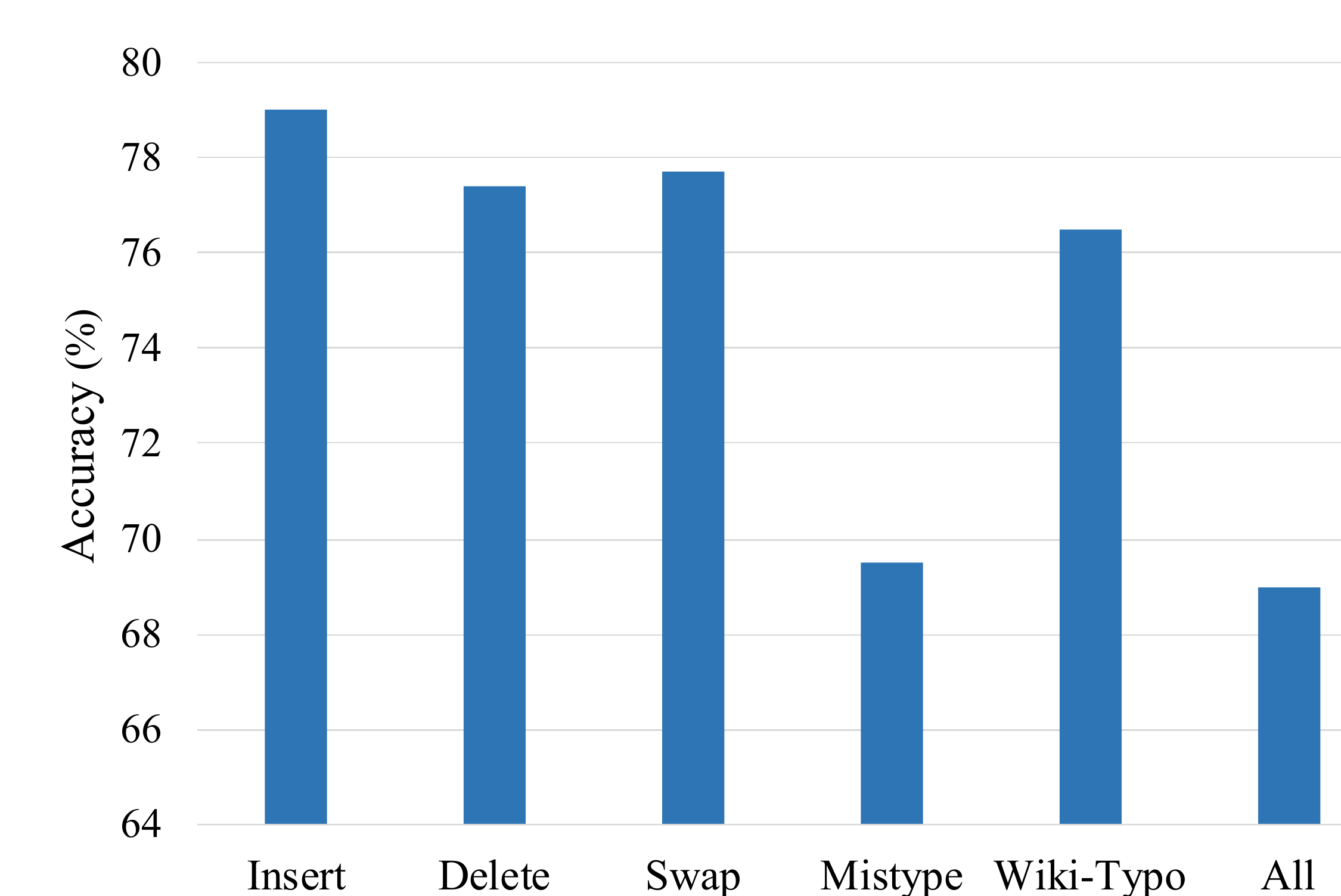}
\caption{Adversarial Typos examples via different types. Left to right: insert, delete, swap, mistype, wiki-typo (common typos by human, including pronunciation and replace-w), all (combines all types of typos).} \label{fig:kinds}
\end{figure}

Next, we further explore the fine-grained influence of the six kinds of word modifications (i.e., ``insertion'', ``deletion'', ``swap'', ``mistype'', ``pronounce'' and ``replace-w'') in Fig. \ref{fig:kinds}. 
Insertion modification has the minimum influence, because sub-words tokenization of BERT would not change much in some cases, such as ``apple'' $\rightarrow$ ``applee''. 
Instead, mistype because of fast typing on the keyboard  hurts the performance most.
The main reason is mistype can generate some uncommon samples, such as ``own'' $\rightarrow$ ``0wn'' or ``9wn''.

\paragraph{Comparison between human and machine.}








To investigate how well humans can read our typo-injected text, we conducted human evaluation. 
As studied in \cite{typo,mt_adv}, some typos are not easily recognized by humans.
We invited 10 persons to read and detect typos in 100 examples used in Table~\ref{tab:sst_results_dev}, then we count the number of sentences where any typos are detected.
The detection would be easy if they spend a long time, so we allocated only three seconds for each sentence.
Table~\ref{tab:human_read} shows the results with $K=1\sim5$, and the scores are the averaged detected counts. Interestingly, the model is sensitive to the ``max-grad'' setting,  humans, instead, are less sensitive---the ``min-grad'' word typos are more eye-catching for humans.
One presumable reason is that the ``min-grad'' setting often injects the typos into some highly frequent words, such as ``the'' and ``it''. People are over familiar with them and their high frequencies in the text increase the chance of being recognized by humans.
Here is an example:

\emph{``{\bf ut}'s a charming and often affecting journey''}\\
where ``it'' is modified to ``ut''.
This looks strange to humans, but the model prediction does not change.

\paragraph{Sensitivity of word segmentation.}
We have observed that humans and the model are sensitive to the different typos (or words).
One explanation of the BERT's sensitivity is that the subword segmentation is sensitive to the typos.
For example, our ``max-grad'' method modifies the keyword ``inspire'' to ``inspird'' in the following sentence:
\begin{itemize}
\item[] \emph{``a subject like this should {\bf inspire} reaction in its audience; the pianist does not.''}
\end{itemize}
As a result, ``inspird'' is split into [ins, \#\#pi, \#\#rd], which is completely different from the original string.
Therefore, one promising direction is to make word segmentation more robust to character-level modifications.
To verify this assumption, 
we trained two other RNN-based classifiers with the widely-used GloVe embeddings~\citep{pennington2014glove} and character n-gram embeddings~\citep{jmt}.
Table~\ref{tab:transfer} shows the results with ``max-grad'' setting.
Our BERT-based typos also degrade the scores of the RNN-based models, and we can see that the character information makes the model  more robust.

\begin{table}[t]

\begin{center}

\begin{tabular}{r|c|c||c}

$K$ & GloVe+char & GloVe & BERT \\ \hline
0 & 85.7 & 85.4 & 92.9 \\ \hdashline
1 & 79.8 & 79.0 & 70.3  \\
2 & 77.5 & 77.2 & 62.3  \\ 
3 & 76.7 & 76.6 & 57.5  \\
4 & 76.3 & 75.9 & 54.1  \\
5 & 76.4 & 75.8 & 53.6  \\ \hline

\end{tabular}
\end{center}
\caption{Transferability of the SST results.}
\label{tab:transfer}

\end{table}

\subsection{Question answering}

\begin{table}[t]

\begin{center}

\begin{tabular}{r|c|c|c}

$K$ & Max-grad & Min-grad & Random \\ \hline
0 & \multicolumn{3}{c}{80.4, 88.2} \\ \hdashline
1 & 23.6, 45.9 & 41.8, 67.6 & 35.6, 60.5 \\ \hline

\end{tabular}
\end{center}
\caption{SQuAD v1.1 results (EM, F1).}
\label{tab:squad_results_dev}

\end{table}

We work on the SQuAD v1.1 benchmark \cite{DBLPRajpurkarZLL16}. The dataset is randomly partitioned into a training set (80\%), a development set (10\%), and a blinded test set (10\%). Evaluation on the SQuAD dataset consists of two metrics: the exact match score (EM) and F1 score.
For this task, we  inject the typos into the questions.

Table~\ref{tab:squad_results_dev} also reports the results when we cast typos in informative, uninformative or random words. 
It is surprising that even a single typo is able to decrease the QA model performance dramatically (therefore, we did not increase the typo size $K$ anymore). Comparing this QA performance and that in sentiment analysis task in  Table~\ref{tab:sst_results_dev}, we notice that the BERT-based QA system is much more brittle than a BERT-based sentiment classifier. It means that the robustness of a NLP system depends on the learning algorithm as well as the task.

\section{Conclusion}

This paper has investigated how the state-of-the-art model, BERT, is robust or brittle to keyboard typos.
Our experimental results show the different sensitivities to different types of words, and suggest the necessity of considering the robustness of the neural models.
We will release our code to reproduce our results, and in future work, we will consider how to make subword-based models more robust to human typos in NLP tasks.

\bibliography{acl2020.bib}
\bibliographystyle{acl_natbib}

\end{document}